\documentclass[conference]{IEEEtran}
\IEEEoverridecommandlockouts
\usepackage{cite}
\usepackage{amsmath,amssymb,amsfonts}
\usepackage{algorithmic}
\usepackage{graphicx}
\usepackage{textcomp}
\usepackage{xcolor}
\def\BibTeX{{\rm B\kern-.05em{\sc i\kern-.025em b}\kern-.08em
    T\kern-.1667em\lower.7ex\hbox{E}\kern-.125emX}}

\begin{document}

\title{LLM-Powered AI Agent Systems and Their Applications in Industry\\
%{\footnotesize \textsuperscript{*}Note: Sub-titles are not captured in Xplore and should not be used} \thanks{Identify applicable funding agency here. If none, delete this.}
}

\author{\IEEEauthorblockN{ Guannan Liang}
\IEEEauthorblockA{ \textit{Independent AI researcher} \\
guannan.liang@yahoo.com
}
\and
\IEEEauthorblockN{Qianqian Tong}
\IEEEauthorblockA{\textit{Computer Science Department} \\
\textit{UNC Greensboro, NC}\\
q\_tong@uncg.edu}
}

\maketitle
\IEEEpubid{979-8-3315-2508-8/25/\$31.00~\copyright2025 IEEE}
\IEEEpubidadjcol

\begin{abstract}
The emergence of Large Language Models (LLMs) has reshaped agent systems. Unlike traditional rule-based agents with limited task scope, LLM-powered agents offer greater flexibility, cross-domain reasoning, and natural language interaction. Moreover, with the integration of multi-modal LLMs, current agent systems are highly capable of processing diverse data modalities, including text, images, audio, and structured tabular data, enabling richer and more adaptive real-world behavior. This paper comprehensively examines the evolution of agent systems from the pre-LLM era to current LLM-powered architectures. We categorize agent systems into software-based, physical, and adaptive hybrid systems, highlighting applications across customer service, software development, manufacturing automation, personalized education, financial trading, and healthcare. We further discuss the primary challenges posed by LLM-powered agents, including high inference latency, output uncertainty, lack of evaluation metrics, and security vulnerabilities, and propose potential solutions to mitigate these concerns. 
\end{abstract}

\begin{IEEEkeywords}
AI Agent, LLMs
\end{IEEEkeywords}

\vspace{-0.3cm}
\section{Introduction}

An \textit{agent} is an autonomous entity capable of perceiving its environment and taking actions to achieve specific goals.
When multiple agents engage in coordination or competition within a shared environment,
they form an \textit{agent system}\cite{oliveira1999multi}. Artificial Intelligence (AI) techniques enable the development of AI agent systems, which integrates perception, reasoning, learning, and action to behave intelligently in a dynamic environment \cite{durante2024agent}.

Recent progress in large language models (LLMs) has significantly changed the AI agent system, driving advances in automation and human-AI collaboration \cite{masterman2024landscape,xie2024large,dorri2018multi,li2024personal,huang2024understanding,guo2024large,li2024survey}.
Compared to traditional agent systems, which mainly relied on task-specific rules \cite{van1978mycin,buchanan1981dendral} or reinforcement learning (RL)\cite{gronauer2022multi,nguyen2020deep,hernandez2019survey,arulkumaran2017deep}, LLM-powered AI agent system provides significantly more adaptability in dynamic and open environments.
Agents can process and generate insights from diverse data modalities, including text, images, audio, and structured tabular data. As a result, current agent system demonstrates the ability to generalize to new tasks, produce contextually rich responses, and enable more natural human-AI interaction.

\begin{figure*}[t]
    \centering
    \includegraphics[width=0.5\textwidth]{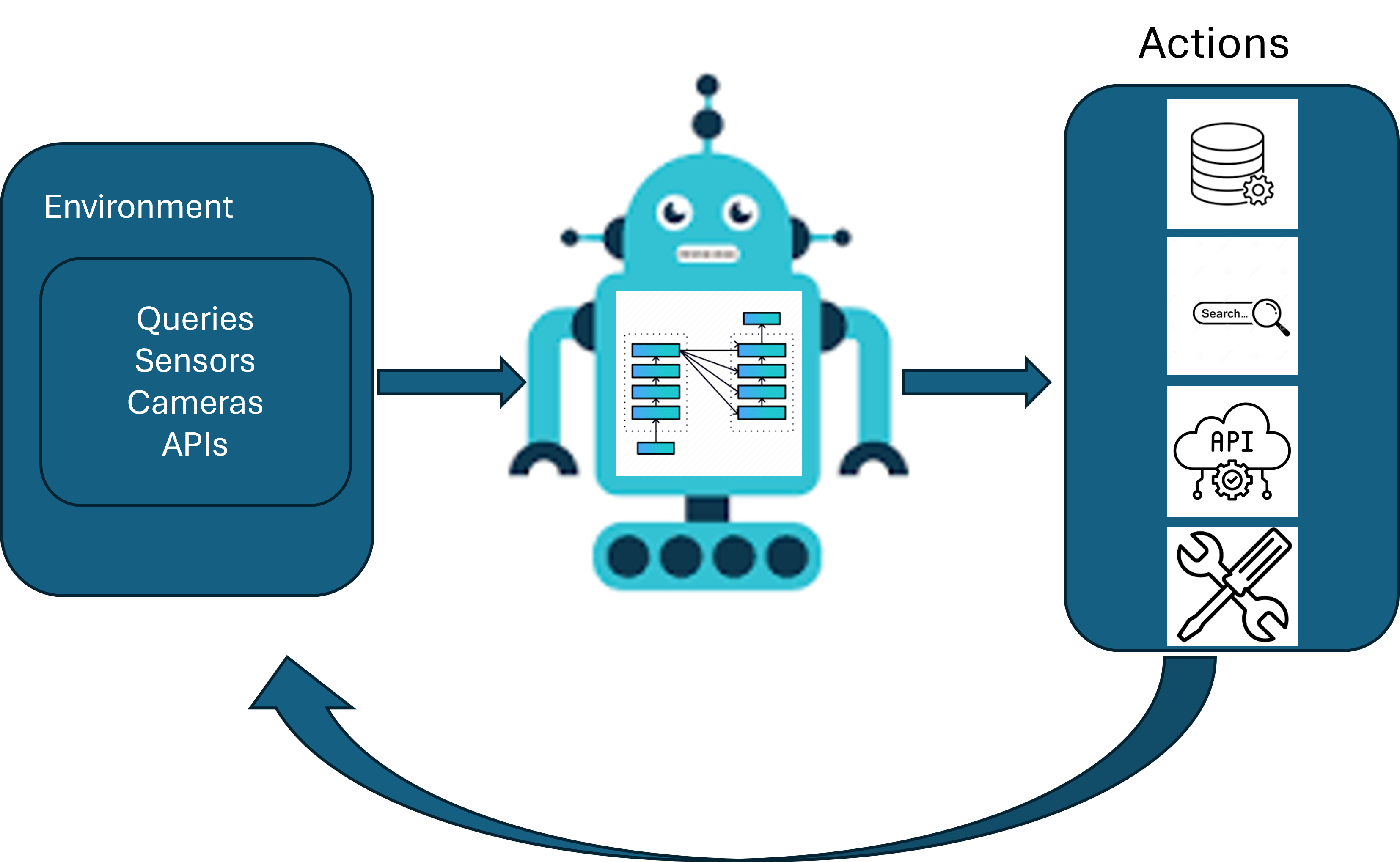}
    \caption{LLM-Powered AI Agent System.}
    \label{fig1:genai_platform}
\end{figure*}

In the LLM era, people often confuse AI agent systems with AI models. To clarify the distinction between the two fundamental concepts in AI, it is crucial to establish precise definitions. An \textit{agent} represents a comprehensive architecture that includes environmental perception, autonomous decision-making, and goal-directed action execution \cite{wooldridge1995intelligent}. Specifically, an \textit{AI agent} is characterized as a self-contained computational entity that: (1) continuously perceives and interprets its environment through various input modalities, (2) processes these perceptions through cognitive functions to make context-aware decisions, and (3) executes appropriate actions to achieve predefined objectives. In contrast, an \textit{AI model} constitutes a specialized computational component that performs specific patterns recognition or data transformation tasks, serving as a functional building block within larger systems. The fundamental distinctions lie in the agent's autonomous capability to initiate and execute actions within its environment. In the LLM paradigm, AI agent systems (Fig.\ref{fig1:genai_platform}) emerge through the systematic integration of multiple AI models with decision-making frameworks, interactive interfaces and automated control mechanisms, thereby creating sophisticated, goal-oriented AI entities.

To better understand the design space of AI agent systems, it is useful to explore their major categories. Agent systems can be categorized according to their mode of interaction with the environment, which determines their operational domain and capabilities. Broadly, we define three major types of agent systems: Software-Based Agents, Physical Agents, Adaptive and Hybrid Agents.

Software-Based Agents (Sandbox Environment) operate entirely in digital environments and interact with users, applications, or online data sources. They do not have a physical presence, but can influence the world through digital means such as APIs, databases, Internet access, and simulated environments.
Here are some of its industry use cases:  LLM-powered Chatbots and virtual assistants - such as ChatGPT\cite{openai2024chatgpt}, Claude\cite{anthropic2024claude}, Gemini\cite{google2024gemini}, DeepSeek\cite{deepseek2024}
- as well as automated financial trading agents \cite{wang2024llmfactor,fatouros2024can,lopez2023can,li2023tradinggpt,zhang2024multimodal}.

In contrast to software-based agents that operate solely in digital environments, physical agents are embodied systems that perceive and act in the real world. Physical agents, which operate in sensor-based environments, interact with the physical world using sensors, actuators, and robotics. 
They interact with the physical world using sensors (such as cameras, LiDAR, and microphones) to perceive their environment, and actuators (including motors, wheels, and robotic arms) to perform actions.
An application of physical agents is in smart manufacturing \cite{keskin2025llm}.

\newpage
By combining the capabilities of software-based and physical agents, hybrid agents emerge as a powerful class of systems that enable seamless integration with the real world. Adaptive and Hybrid Agents (Real-World Integration) operate in a feedback-driven environment, continuously learning from both digital and physical interactions by processing multi-modal data such as text, images, voice, and sensor inputs, and adapting their decision-making over time. Here are some of its industry use cases:
AI-driven traffic management optimizing real-time road congestion \cite{dikshit2023use},
Healthcare AI assistants: agents that monitor patient data, recommend treatments and interact with doctors\cite{kim2024mdagents,wei2024medaide,mukherjee2024polaris,mehandru2024evaluating},
AI-powered predictive maintenance systems combining software analytics with sensor data\cite{mazumdarai,abbas2024ai},
AI-powered supply chain management that optimize logistics by integrating real-world shipment tracking with digital AI forecasting\cite{shobhana2024ai}.

Most existing work or surveys on agent systems fall into two major directions: one focusing on theoretical foundations such as agent modeling and multi-agent coordination, and the other on practical frameworks such as reinforcement learning and system implementation \cite{masterman2024landscape,durante2024agent,xie2024large,dorri2018multi,li2024personal,huang2024understanding,guo2024large,li2024survey,gronauer2022multi,nguyen2020deep,hernandez2019survey,arulkumaran2017deep}. However, the applications of LLM-powerd AI agent systems in real-world industry settings remain relatively underexplored, despite growing interest and potential impact.
This paper aims to provide some insights from an industry-focused perspective on the development and categorization of LLM-powered agent systems.

\section{Agent Systems Overview}
\subsection{ Agent Systems Before LLM era}

Before the emergence of LLM-powered agents,
traditional agents were typically based on rule-based logic, search, planning, or RL, and were often designed for narrow and task-specific domains.
These systems were effective in structured environments; however, when dealing with unstructured data such as natural language or images, or when transferred to new environments, they suffered from poor generalization, lacked adaptability, and were not capable of natural human-AI interaction.

Rule-based agent systems were among the earliest forms of agent, relying on predefined rules and decision trees to perform reasoning. Expert systems such as MYCIN \cite{van1978mycin} for medical diagnosis and DENDRAL \cite{buchanan1981dendral} for chemical analysis use  structured if-then rules to make domain-specific decisions. Throughout the 1980s and 1990s, business rule engines, decision trees, and logic programming systems had been widely used across industries such as healthcare, finance and manufacturing\cite{davenport2005automated}. The introduction of Multi-Agent Systems in the 1990s enabled distributed decision-making, allowing agents to interact with each other to solve complex problems such as supply chain management and automated trading\cite{dorri2018multi}.

Following rule-based agents, RL emerged as a powerful framework for training agents to make sequential decisions by interacting with their environment \cite{gronauer2022multi,nguyen2020deep,hernandez2019survey,arulkumaran2017deep}. RL agents learn optimal behaviors through a process of trial and error, driven by the maximization of cumulative rewards. Unlike predefined rule, RL agents autonomously discover strategies by receiving feedback in the form of rewards or penalties. Key algorithms, such as Q-learning\cite{clifton2020q,kumar2020conservative}, Deep Q-Networks (DQN)\cite{van2016deep,hafiz2022survey}, and Policy Gradient Methods\cite{sutton1999policy}, enabled RL agents to master complex tasks, including game playing, robotic control, and dynamic resource management. In particular, systems such as AlphaGo \cite{silver2017mastering} demonstrated the power of RL by defeating human champions in the game of Go, showcasing the potential of agents to surpass human expertise in structured environments. However, RL agents are highly task-specific and often fail to generalize well, requiring retraining even for minor environment changes — unlike LLMs, which can adapt to tasks with zero-shot or few-shot prompting.

\subsection{LLM-powered Agent System}

LLM-powered agents leverage LLMs \cite{naveed2023comprehensive} and multi-modal foundation models\cite{lin2025survey} to enable flexible and adaptive decision-making. These agents can process text, images and audio, making them applicable across diverse industries, including healthcare\cite{wang2025survey}, finance\cite{ding2024large}, and manufacturing\cite{li2024large,garcia2024framework,zhao2024large}. Unlike rule-based and RL agents, LLM-powered agents do not rely on predefined decision trees and expensive explorations, enabling them to generalize to new and evolving tasks.

%These agents are capable of performing a wide range of tasks, including natural language understanding, autonomous problem-solving, real-time decision-making, and human-like interaction. In healthcare, LLM-powered agents assist in medical diagnosis, and personalized treatment recommendations by analyzing patient records and medical literature. In finance, they play a crucial role in  risk assessment and algorithmic trading, leveraging real-time financial news analysis. In manufacturing, these agents optimize predictive maintenance, supply chain management, and quality control, reducing operational costs and improving efficiency.

These agents are capable of a wide range of advanced functions, including natural language understanding, autonomous problem-solving, planning, reasoning, and human-like interaction. More importantly, LLM-powered agents significantly enhance human-AI collaboration by enabling context-aware dialogue, intelligent virtual assistance, and real-time decision support. Their ability to process multi-modal inputs — such as text, images, and voice — allows them to operate effectively in diverse tasks, including visual data interpretation, multi-turn conversations, and voice-guided workflows. This versatility makes them valuable tools in a complex, data-rich environment.

Alongside their strengths, LLM-powered agents also face several primary challenges, including high inference latency, the lack of standardized benchmarks and evaluation metrics, and privacy concerns. We will discuss these challenges in Section IV.

\subsection{Architecture of LLM-Powered Agent System}

Theoretically, a LLM-powered agent system integrates several critical components: 
(1) At its core, the \textbf{LLM} serves as the cognitive engine, responsible for high-level reasoning, planning, and natural language understanding. Surrounding this core are supporting modules that extend its capabilities: 
(2) \textbf{Tool utilization}, facilitated by techniques like Multi-Context Prompting (MCP)\cite{anthropic2024mcp,singh2025survey}, allows the agent to dynamically invoke APIs, databases, or third-party models to accomplish specialized tasks;
(3) \textbf{Memory}, typically implemented via Retrieval-Augmented Generation (RAG)\cite{lewis2020retrieval,gao2023retrieval}, ensures the agent can access external knowledge and avoid hallucination \cite{maynez2020faithfulness,ji2023survey};
(4) \textbf{Environmental sensing}, which is achieved through multi-modal inputs—such as text, images, speech, or sensor data—captured by models or devices like cameras and IoT hardware, enabling the agent to perceive and react to its surroundings; 
(5) An essential auxiliary layer is the \textbf{guardrail} mechanism\cite{inan2023llama,dong2024safeguarding,dong2024building}, which filters both inputs and outputs for safety, compliance, and task relevance, thereby ensuring trustworthy and reliable behavior in real-world deployments.

\begin{figure*}[t]
    \centering
    \includegraphics[width=1\textwidth]{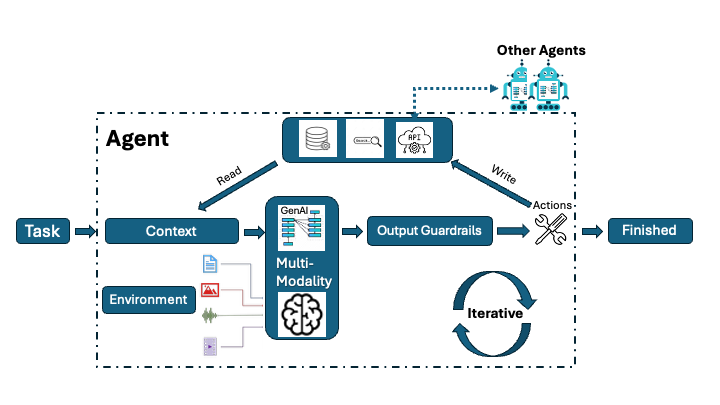}
    \caption{Architecture of LLM-Powered Agent System.}
    \label{fig2:genai_platform}
\end{figure*}

In practice, we summarize the LLM-powered agent system in the framework as shown in Figure \ref{fig2:genai_platform}, where each component ensures contextual awareness, decision efficiency, and reliable execution. 
The LLM agent system starts with a Task Input, which defines the objective and provides textual or structured instructions that the agent must process. 
Once the task is identified, the agent performs Context Augmentation, leveraging external knowledge sources such as databases, search engines, and API calls to other agents. This step ensures that the agent can ground its decisions in relevant and up-to-date information rather than relying only on its pre-trained knowledge. 
The system then enters a Decision and Planning Phase, where the LLM model generates a structured response based on environmental context, retrieved knowledge, and multi-modal data. However, since GenAI output can be highly variable and sometimes unstructured, an Output Guardrail Mechanism ensures compliance with predefined formats, validation rules, and industry-specific constraints before final execution. The final Action Execution step translates structured outputs into real-world commands, whether it is interacting with software, triggering automation scripts, or instructing robotic systems. The system operates iteratively, continuously sensing the environment, adjusting plans, and refining actions until the goal is successfully achieved.

\section{Industry Applications}

Previous surveys \cite{masterman2024landscape,durante2024agent,xie2024large,dorri2018multi,li2024personal,huang2024understanding,guo2024large,li2024survey} on LLM agents mainly focused on theoretical foundations, however, system design and applications across industries still remains underexplored. In this work, we categorize industry applications into the following key domains.

\subsection{Chatbot: Live Customer Service}
LLM-powered Chatbots have revolutionized customer interaction by providing dynamic, context-aware, and natural language responses. Unlike traditional rule-based systems, modern Chatbots can understand complex queries and significantly improve the accuracy of responses through LLMs \cite{abedu2024llm,kumar2023large,dam2024complete,sufi2025just}. Chatbots are widely used in customer support, marketing automation, and interactive user interfaces. For example, LLM-powered Chatbot systems, such as 
ChatGPT\cite{openai2024chatgpt}, Claude\cite{anthropic2024claude}, Gemini\cite{google2024gemini}, DeepSeek\cite{deepseek2024},  have been integrated into e-Commerce websites to assist users with product recommendations, query resolution, and after-sales support\cite{iyer202413}. 

%Prominent examples of such systems include ChatGPT, which powers customer service and knowledge base interactions for businesses using the API or through integrations in platforms like Shopify and Microsoft Copilot; Google Gemini, which has been embedded into Google Workspace to enhance email drafting, document summarization, and meeting assistance; and Anthropic's Claude, known for its alignment with safety and helpfulness, used in sensitive customer-facing applications such as legal support and healthcare triage. DeepSeek has also been adopted by local enterprises for multilingual support and real-time translation in global customer service pipelines. These advanced LLM chatbots not only reduce operational cost by automating repetitive interactions but also enhance user satisfaction by delivering more intelligent, engaging, and contextually relevant conversations.

\subsection{Software Development}
LLM-based coding assistants are transforming the software development process by automating code generation, debugging, and documentation\cite{jin2024llms,lyu2024automatic}. These agents leverage advanced natural language understanding to translate human instructions into executable code, thereby reducing manual effort and enhancing productivity. For instance, the use of automated code synthesis agents, as discussed in previous literature, has significantly lowered the barrier for non-expert developers to implement complex algorithms. In cybersecurity, LLM-based defense and attack agents can proactively detect vulnerabilities and simulate potential exploits, enhancing system robustness\cite{hassanin2024comprehensive,kasri2025vulnerability}. In addition, industry-focused coding agents, like Github Copilot\cite{githubcopilot}, Cursor\cite{cursor2024}, help developers by predicting code snippets and suggesting improvements, making software engineering more efficient.

\subsection{Manufacturing Automation}
In the manufacturing sector, LLM-powered robots facilitate automated decision-making and precision control\cite{li2024large,garcia2024framework,zhao2024large}. They are utilized for automating tasks such as product design, quality control, and supply chain management by interpreting complex instructions and extracting insights from extensive datasets. Additionally, they play a role in enhancing robot control systems, leading to more innovative and efficient manufacturing practices. For example, the LLM-powered agent can significantly advance the development of human-centric smart manufacturing by facilitating the integration of human–robot interaction and collaboration\cite{keskin2025llm}.

\subsection{Personalized Education}
LLM-powered agents are reshaping personalized education by providing dynamic, context-aware support for both learners and educators across diverse educational tasks \cite{chu2025llm,sharma2025role}. These agents serve two primary functions: as teaching assistants that automate instructional planning, resource recommendation, classroom simulation, and feedback generation \cite{xu2024eduagent,jin2024teachtune,mower2025khwarizmi} and as personalized learning assistants that adaptively support student learning paths\cite{xu2024eduagent,yang2024content}. 
By leveraging memory, tool-use, and planning capabilities, LLM-powered agents can track student progress, identify learning gaps, and dynamically generate customized exercises and feedback.  
Additionally, LLM-powered agents are being deployed in domain-specific education, such as mathematics \cite{yan2025mathagent} and science \cite{wang2023newton,m2024augmenting}, where they guide students through complex reasoning tasks. 
This multifaceted support not only enhances learner engagement and understanding, but also reduces educators' workload, marking LLM-powered agents as a foundational component in next-generation educational ecosystem.

\subsection{Healthcare}

In healthcare, LLM-powered agents facilitate patient interaction, medical record analysis, and clinical decision support. These agents can extract key information from patient histories, synthesize medical knowledge, and generate diagnostic recommendations\cite{wang2025survey}. For instance, conversational healthcare agents assist doctors by summarizing patient reports and suggesting evidence-based treatments. LLM-powered agents such as MDAgents\cite{kim2024mdagents}, MedAide\cite{wei2024medaide}, and Polaris\cite{mukherjee2024polaris} demonstrate how they can simulate doctor-patient interactions, reduce cognitive biases, and coordinate multi-agent workflows across clinical stages including diagnosis, medication, and follow-up care. Moreover, LLM-based diagnostic assistants can interact with tools and databases\cite{mehandru2024evaluating}, which improve precision by correlating symptoms with medical databases, offering clinicians a more comprehensive view of potential diagnoses. Such applications improve care quality while minimizing administrative workload.

\subsection{Financial Trading}

LLM-powered trading agents are emerging as a transformative technology in financial markets, leveraging large-scale language models to process unstructured data, generate trading insights, and make informed investment decisions\cite{ding2024large}. These agents are typically designed to serve two roles: LLM agents as traders \cite{wang2024llmfactor,fatouros2024can,lopez2023can,li2023tradinggpt,zhang2024multimodal}, where the model directly outputs buy-hold-sell signals, and LLM agents as alpha miners \cite{wang2024quantagent,wang2023alpha}, where the model discovers new predictive features for quantitative strategies. Agents excel at integrating multiple data modalities: textual (e.g. financial news, reports), numerical (e.g. price time series), visual (e.g. trading charts), and simulated data to generate predictions or trading actions. These agent systems demonstrate strong performance in backtesting. Despite promising results, challenges remain in latency, generalization, and system integration, especially for high-frequency trading and real-time decision-making environments.

%\subsection{Simulation Environment}
%LLM agents are increasingly utilized in simulation environments to model complex systems and predict dynamic changes\cite{gurcan2024llm,duvvuru2025llm,gao2024large}.

\section{Challenges}

\subsection{High Inference Latency}
One of the most significant challenges faced by LLM-powered agents in real-world applications is high inference latency, which directly results in high operational costs and computational inefficiencies. Due to their massive parameter size, LLMs (such as GPT-4) introduce significant inference overhead and
long execution latency\cite{xu2024deploying}. This results in long response times, especially when deployed for tasks requiring real-time interaction, like customer support, financial trading, or industrial monitoring.

The issue becomes even more pronounced in environments where quick decision-making is critical. For example, in healthcare settings where agents are expected to provide rapid diagnostic assistance, high latency can compromise patient care. In financial trading, delayed responses can result in missed market opportunities or significant financial loss. High latency not only affects the user experience but also increases infrastructure costs, as maintaining low-latency responses often necessitates deploying multiple high-performance GPUs or TPUs. Moreover, as LLM models scale up, inference times become increasingly impractical for applications requiring low-latency processing.

To address high inference latency in LLM-powered agents, a combination of model compression and efficient deployment strategies is essential. Compressed models with quantization\cite{frantar2022gptq,lin2024awq,liu2023llm,yao2022zeroquant,xiao2023smoothquant}, pruning\cite{ma2023llm,frantar2023sparsegpt,sun2023simple}, and knowledge distillation \cite{timiryasov2023baby,gu2023minillm,hsieh2023distilling} reduce model size and speed up processing without a significant loss of accuracy. Efficient deployment practices, including optimizations for computation \cite{lu2020hardware,jang2019mnnfast,bai2023transformer,zeng2024flightllm,sheng2023flexgen} and memory \cite{alizadeh2024llm,kwon2023efficient} further mitigate latency issues. Additionally, deploying LLM agents closer to users via edge computing and leveraging hardware accelerators like TPUs enhance real-time responsiveness. Implementing caching mechanisms for repetitive tasks and using adaptive sampling to minimize unnecessary full-model inference also contribute to reducing processing time and operational costs, making LLM agents more practical for real-time applications.

\subsection{Uncertainty of LLM Output}
LLM agents often face the challenge of producing uncertain or unreliable outputs\cite{shorinwa2024survey}, or hallucination\cite{huang2025survey}. Since LLMs are trained on vast and diverse datasets, they may generate responses that are contextually inaccurate, biased, or even factually incorrect. This uncertainty is problematic in high-stakes applications, such as legal document analysis, medical diagnosis, or autonomous systems, where incorrect outputs can lead to significant consequences. Additionally, LLMs can generate hallucinations, plausible-sounding but entirely false information, which can undermine user trust and the reliability of the system. This uncertainty makes it challenging to integrate LLM agents into mission-critical environments where accuracy and consistency are paramount.

To address the uncertainty of LLM output, it is essential to integrate a guardrail layer after the output layer, acting as a post-processing validation mechanism\cite{inan2023llama,dong2024safeguarding,dong2024building}. This guarding layer can employ techniques such as factual consistency checks, out-of-distribution detection, and context verification to filter and refine the generated responses. Additionally, leveraging ensemble methods where multiple LLM instances provide responses and voting mechanisms select the most consistent output can enhance reliability \cite{chen2025harnessing,naik2024probabilistic,yang2023one}. Integrating external knowledge bases, obtained by Retrieval-Augmented Generation (RAG) \cite{gao2023retrieval} and external tools\cite{shen2024llm,singh2025survey}, to cross-check facts can also reduce misinformation risks. In high-stakes applications, incorporating a human-in-the-loop system ensures that sensitive outputs undergo manual validation before deployment. These strategies collectively improve the robustness of LLM agents, minimizing accumulated errors and maintaining user trust.

\subsection{Lack of Benchmarks and Evaluation Metrics}

One of the critical challenges in the development and deployment of LLM-powered agent systems is the lack of standardized benchmarks and evaluation metrics\cite{kapoor2024ai}. Although there are numerous studies and metrics designed to evaluate the performance of LLMs themselves, these metrics often fall short when applied to complex agent systems that involve decision-making, multi-modal processing, and human-AI interactions\cite{xie2024large}. LLM-based agents go beyond generating coherent text—they perform tasks such as planning, interacting with other systems, and adapting to dynamic environments. As a result, evaluating their performance requires a comprehensive approach that considers not only linguistic accuracy but also task success rate, adaptability, context awareness, and human satisfaction. The absence of universally accepted benchmarks makes it challenging to compare different systems, track progress, and ensure reliability in practical applications.

Hence, developing domain-specific benchmarks and multi-dimensional evaluation frameworks that account for the diverse functionalities of LLM-powered agents is necessary. One promising approach is to create task-oriented metrics that assess performance based on goal achievement and interaction quality, such as task success rate, dialogue coherence, and response accuracy \cite{wu2023smartplay,mialon2023gaia,koh2024visualwebarena,lu2024weblinx,xie2024travelplanner}. Additional, employing human-centric evaluation methods — such as user satisfaction surveys and feedback — can capture qualitative aspects that automated metrics might miss \cite{yan2023gpt}. Another strategy is to develop simulation environments that mimic real-world tasks, allowing for controlled and reproducible testing of agent performance. Benchmarking competitions and shared datasets curated for multi-agent interactions \cite{guo2024large} and real-world scenarios \cite{geng2025realm} can also help establish community standards. By combining quantitative metrics with qualitative assessments, the evaluation of LLM-powered agents can become more comprehensive, enabling consistent comparisons and fostering innovation.

\subsection{Security and Privacy Concerns}
Security and privacy are significant challenges when deploying LLM-powered agent systems due to their susceptibility to attacks and data leakage. One of the most critical issues is the jailbreak of AI agent systems, where adversaries manipulate the input prompts to bypass safety mechanisms, leading the agent to generate harmful or unethical content\cite{gu2024agent,dong2024jailbreaking,liu2024jailjudge}. Such vulnerabilities can be exploited to spread misinformation, generate malicious code, or perform unauthorized actions. Additionally, LLM agents often process sensitive user data, posing risks of unintentional data leakage. Attackers can use prompt injection attacks to trick the model into revealing proprietary or personal information \cite{he2024emerged}. Furthermore, model inversion attacks may reconstruct training data from the model's responses, violating data privacy regulations. These security loopholes can result in significant harm, especially in industries where data confidentiality and system integrity are paramount, such as finance, healthcare, and critical infrastructure.

To address security and privacy concerns, it is essential to implement multi-layered defense mechanisms throughout the LLM-powered agent system. First, incorporating a guarding layer that actively filters and validates user inputs can help detect and block potential jailbreak attempts \cite{dong2024safeguarding}. Utilizing adversarial training techniques, where the model is exposed to harmful prompts during training, can increase robustness against prompt injection \cite{kumar2023certifying,xhonneux2024efficient}. Moreover, employing differential privacy techniques ensures that the model's responses do not inadvertently reveal individual data points from the training set \cite{behnia2022ew,singh2024whispered}. Deploying content moderation pipelines \cite{franco2024integrating} and using response validation mechanisms \cite{kulsum2024case} can further safeguard against generating inappropriate or harmful outputs.  By integrating these proactive measures, LLM-powered agent systems can be built significantly more secure and privacy-compliant.

\section{Conclusion}
This paper studies LLM-powered AI agent systems, tracing their evolution from rule-based and reinforcement learning frameworks to modern LLM-driven architectures. By examining both historical and contemporary developments, we provide a structured understanding of how LLMs and multi-modal AI techniques are shaping next-generation intelligent agents. We then demonstrate how LLM-powered agents are revolutionizing diverse industries by enabling automation, intelligent decision-making, and enhanced human-AI collaboration.
Along with the strong capabilities, LLM-powered agents still face several challenges, such as high inference latency, output uncertainty, inadequate evaluation metrics, and security vulnerabilities. Addressing these issues requires a combination of model optimization, efficient deployment strategies, robust evaluation frameworks, and multi-layered security protocols. As LLM-powered agents continue to evolve, it is crucial to develop solutions that enhance their scalability, reliability, and ethical compliance. Future research should focus on creating more adaptive and context-aware agents that seamlessly integrate into complex industrial ecosystems. By addressing existing challenges, the next generation of intelligent agent systems can achieve broader and more reliable real-world applications.

\bibliographystyle{IEEEtran}
\bibliography{references}
\end{document}